\documentclass{article}

\usepackage{arxiv}

\usepackage[utf8]{inputenc} 
\usepackage[T1]{fontenc}    
\usepackage{hyperref}       
\usepackage{url}            
\usepackage{booktabs}       
\usepackage{amsfonts}       
\usepackage{nicefrac}       
\usepackage{microtype}      
\usepackage{lipsum}
\usepackage{graphicx}
\usepackage{natbib} 
\usepackage{array}
\graphicspath{ {./images/} }

\title{An LLM-based Delphi Study to Predict GenAI Evolution}

\author{
 Francesco Bertolotti \\
  School of Industrial Engineering\\
  LIUC - Università Cattaneo\\
  Castellanza (Italy), 21053 \\
  \texttt{fbertolotti@liuc.it} \\
   \And
 Luca Mari \\
  School of Industrial Engineering\\
  LIUC - Università Cattaneo\\
  Castellanza (Italy), 21053 \\
  \texttt{fbertolotti@liuc.it} 
}

\begin{document}
\maketitle
\begin{abstract}

\end{abstract}

Predicting the future trajectory of complex and rapidly evolving systems remains a significant challenge, particularly in domains where data is scarce or unreliable. This study introduces a novel approach to qualitative forecasting by leveraging Large Language Models to conduct Delphi studies. The methodology was applied to explore the future evolution of Generative Artificial Intelligence, revealing insights into key factors such as geopolitical tensions, economic disparities, regulatory frameworks, and ethical considerations. The results highlight how LLM-based Delphi studies can facilitate structured scenario analysis, capturing diverse perspectives while mitigating issues such as respondent fatigue. However, limitations emerge in terms of knowledge cutoffs, inherent biases, and sensitivity to initial conditions. While the approach provides an innovative means for structured foresight, this method could be also considered as a novel form of reasoning. further research is needed to refine its ability to manage heterogeneity, improve reliability, and integrate external data sources.


\section{Introduction}

Performing predictions of complex and inherently unpredictable systems is a critical and challenging task \citep{batty2005modelling}, which modern technology has only partially addressed \citep{fan2020long}. In situations where data are available, new methodologies have emerged \citep{ghadami2022data}, enabling predictive capabilities even in difficult contexts, typically through the use of machine learning techniques \citep{pathak2018hybrid}.

Nevertheless, there remains a broad set of relevant situations in which making predictions is crucial \citep{bertolotti2024prediction}, yet no data are available, rendering modern data-driven methodologies inapplicable \citep{fletcher2020missing, bertolotti2024balancing}. For example, when data are unavailable but the system is not overly complex, it is possible to develop an intuition about its behavior and translate this understanding into a mathematical formulation, thereby generating a dynamical model that describes the system's behavior and can be used to hypothesize its future states \citep{lotka1920analytical, volterra1928variations, kermack1927contribution}. However, some systems exhibit a level of complexity that prevents the formulation of a simple mathematical model \citep{petty2018modeling, occa2022understanding}. Moreover, certain systems not only change over time but also alter the very mechanisms governing their change \citep{holland1992complex}. In such cases, when conditions are unstable, even dynamical models—which capture underlying cause-effect relationships—become ineffective, as they are based on past states \citep{bertolotti2020sensitivity}. Under these circumstances, quantitative prediction becomes increasingly difficult, if not impossible. One potential approach is to qualitatively anticipate future developments through a Delphi study \citep{dalkey1963experimental}. 

A Delphi study is a structured, iterative method used for forecasting \citep{hasson2011enhancing, gallego2014exploring}. It relies on the systematic collection and refinement of expert opinions to reach a consensus on a given topic \citep{lombardi2025increased}. Through multiple rounds of anonymous surveys, experts provide feedback, which is aggregated and analyzed to refine their responses in subsequent iterations, ultimately aiming to achieve consensus among experts on a specific issue \citep{niederberger2021coming}. This method is particularly useful for exploring complex or uncertain issues where empirical data are limited or unavailable \citep{grisham2009delphi}. Thus, in cases where systems are evolving rapidly and significantly, rendering data-driven and causal methods unsuitable—since the generative process producing the data is itself changing over time, making inference on past data no longer valid—a potential solution is to perform a qualitative prediction of the system’s trajectory \citep{Dueker2005Dynamic, zhu2010novel}. The Delphi method relies on subjective judgments, making it susceptible to respondent bias, as expert opinions may be influenced by personal perspectives rather than objective criteria \citep{Hallowell2009Techniques, Ecken2011Desirability}. Even when quantitative data are available, they cannot be effectively integrated, limiting the method’s ability to incorporate empirical evidence into the forecasting process \citep{Tapio2011The, Mitchell2021Editing}. Moreover, and most importantly, Delphi studies typically yield qualitative insights without establishing formal functional relationships between variables, preventing the formulation of precise or numerical predictions \citep{Khamis2022The}. This lack of quantitative rigor makes it unsuitable for contexts requiring precise forecasts \citep{Gan2015Delphi}. Nevertheless, this approach serves as a valuable methodology for addressing long-term predictions involving dynamic systems, which would otherwise be unmanageable through alternative means \citep{petty2018modeling, fletcher2020missing}.

In this paper, we propose a new methodology for conducting Delphi studies, no longer based on expert commentary but instead on text generated by Large Language Models (LLMs) \citep{vaswani2017attention, schoenegger2025ai}. To the best of our knowledge, this is the first work to propose such an approach. The underlying intuition is that a sufficiently intelligent model could, at least in principle, be employed in a brainstorming process as a substitute for a human participant, an assumption similar to that underlying reasoning models \citep{yao2023tree, jaech2024openai}. We do not claim that LLMs possess expertise comparable to that of human experts in a specific domain, although multiple recent studies support this possibility \citep{ullah2024challenges, krishnaveti2024gpt}. Rather, we aim to test a potential methodology that could, in the future, decouple long-term and complex forecasting from the necessity of access to human experts.

In this study, we tested this methodology specifically in the context of the future of generative artificial intelligence (GenAI) \citep{ooi2025potential}. This problem was selected for three main reasons. First, it is highly relevant, and any insights gained from the study could be valuable to practitioners and policymakers in making informed decisions in this field. Second, it is a problem that is far too complex to be modeled using a dynamical system, and while some data are available, they are insufficient to enable a reliable prediction. 

This paper has a threefold purpose. First, it provides a framework for utilizing LLMs to perform predictions in highly complex contexts where data are unavailable and qualitative insights are crucial. In this regard, the study aligns with previous literature on Delphi studies while fundamentally redefining the approach by eliminating the necessity of human participants for this type of analysis. Additionally, it addresses some of the typical limitations of Delphi studies, such as respondent fatigue. Here, the primary constraint is computational capacity (or API costs), meaning there is no inherent limit to the number of open-ended questions that can be presented to participants. Second, it proposes a new potential architecture for reasoning models. More specifically, since the release of GPT-4o in May 2024, models capable of performing reasoning have been the most effective in solving complex problems. In this context, when viewed from a sufficiently broad perspective, the proposed system can be interpreted as a single cognitive entity executing a highly structured form of reasoning—one with a well-established history of success. Finally, the study generates insights into the potential evolution of GenAI, not only serving as an illustrative case study but also addressing a topic of significant relevance. These findings could provide valuable guidance for both scholars and practitioners in shaping their strategic and professional decisions.

The study demonstrates that LLM-based Delphi methods can effectively generate structured foresight on the evolution of GenAI. Key findings highlight the influence of geopolitical tensions, economic disparities, and regulatory frameworks on GenAI development. The methodology enables extensive scenario exploration, revealing potential pathways for GenAI governance, collaboration, and ethical considerations. Results suggest that GenAI’s societal impact depends on interdisciplinary cooperation, cultural adaptation, and economic accessibility. Notably, while LLMs mitigate human biases and fatigue, they introduce knowledge limitations due to training data constraints. Discussions emphasize the potential for GenAI-driven regulatory mechanisms, decentralized governance models, and dynamic ethical frameworks. Also, the study identifies challenges in ensuring knowledge heterogeneity and managing computational costs. Overall, the LLM-based Delphi approach offers a promising yet evolving tool for qualitative forecasting in complex, data-scarce domains.

The paper proceeds as follows. First, the methodology is thoroughly presented, including the Delphi process, the experimental setup, and the implementation details. Then, the results are presented and discussed. Finally, conclusions are drawn.

\section{Methods}

\subsection{Methodology description}

Following the classic Delphi method, the methodology of this work consists of a simple process where some of the steps can be iterated multiple times \citep{dalkey1963experimental}. In this section, each step is described. The idea behind specifying the process here is not only to provide replicability and an explanation of the work but also to make it possible to see the entire Delphi as a single stochastic function that receives a set of open questions, a set of possible roles, and a topic, and returns a prediction.

\paragraph{Agents structure}
Exactly as in a traditional Delphi study, the system was designed to include two types of agents, each with distinct roles.
The first is the organizing agent, which assumes the role typically played by the facilitator in a classical Delphi study. This agent collects participants' responses, reformulates them, generates questions for the questionnaire, and, based on the questionnaire responses, produces new open-ended questions while also determining when consensus has been reached. The organizing agent in our system performs precisely these tasks. The second type of agent is the responding agent, which simulates an expert. This agent is responsible for answering both open-ended and closed-ended questions.

In this implementation, the same LLM was used for both agents, meaning they ultimately differ only in two aspects: (i) the system prompt, which defines their assigned role and serves as the sole mechanism for customizing their behavior; and (ii) the set of possible actions attributed to each class of agents. The organizing agent can collect closed-ended questions and transform them into open-ended ones, and vice versa. In contrast, the responding agents are limited to providing answers, either in textual form or as numerical values, depending on the question type.

\paragraph{Problem definition}
Each problem-solving initially required a problem to be solved. In this case, the definition of the problem does not have specific boundaries, but could follow specific recommendation. First, the method is suited for cases where data-driven tecniques are not available, because sufficient data are present \citep{peffers2005planning}. Second, is suited when it is not possible to get an understanding of the underlying cause-effect relationship, and so it is not possible to get an insight. Third, even when data are available or it is possible to make a formal model, it is a good idea to use it if it is expected the underlying system to change in an observable way during the time range of the prediction. Fourth, the problem could benefit from getting a multiple-perspective solution \citep{linstone1975delphi}. Fifth, whenever it the problem is actually relevant to face in this manner, given that the technique is computationally costly, and it requires time and effort to set. Finally, the Delphi method is not designed for result generalization; rather, its goal is to offer a deep understanding of a complex issue \citep{gordon1994delphi}.

\paragraph{Experts definitions}
Choosing the right Delphi participants is a crucial challenge in the Delphi method, as it directly impacts the quality of the outcomes produced \citep{hasson2011enhancing}. In the classic Delphi methods, the expert can not be created but instead should be selected in a set of available experts with whom there is a contact and that they know they can answer. The actual numerosity is a relevant element. Several researchers have noted that most Delphi studies typically involve between 15 and 20 participants \citep{alarabiat2019delphi}. However, certain studies have engaged a substantially larger pool of experts, with some including several hundred or even thousands of participants \citep{gallego2014exploring}. Some argue that achieving higher levels of reliability necessitates a larger number of participants, as an increase in panel size is associated with enhanced reliability of the results \citep{donohoe2009moving}, but there is not an agreement regarding this point \citep{akins2005stability}.

In our case, we wanted to have an heterogeneity of responsants, which could be considered expert in the field. For this reason, we have assigned roles to the LLMs agents that are used in the experiments. More specifically, each LLM is defined by nationality, educational background, type of experience, the field of experience, and the field of specialization. The specific configuration has been randomly sampled at the beginning of the simulation, with each occurrence of the feature having a specific probability of occurring. This has been designed to guarantee both replicability (which is needed if one consider this methodology as an special case of the Delphi study) and stochasticity (if it is a cognitive architecture). The roles is then passed to them in the system prompt when they answer, no further specification of the behavioir has been implemented.

\paragraph{Open questions definition}
In a Delphi study, open-ended questions should be carefully designed to minimize ambiguity and bias, ensuring a balanced communication structure that aligns with the study’s objectives and the expertise of the participants \citep{linstone1975delphi}, especially that it may lead to broad or ambiguous responses, potentially introducing bias in both expert contributions and the overall results \citep{hsu2019minimizing}. We opted for open-endend questions, in order to maximize the generation of new ideas in the process. Beside that, open-ended questions must be clear and specific enough to guide the LLMs toward relevant insights without overly constraining their responses; the wording should promote depth and critical reflection while avoiding excessive generalization, which could introduce bias. 

In this case, to get control over the experiment and being able to actually rpoduce different results with different outputs, and conversely scientifically assess the quality of the method, what we have done is to generate manually the first open-endend questions and used them as one of the input of the system, for the sake of the functional thinking depicted at the beginning of the methodology description.

\paragraph{Open question erogation}
The open questions are provided by a prompt to the LLM, so that they have the conxtest to being able to answer according to the setting. The specifics of the prompt is possibile to find in the online repository. The open-ended questions are provided and a short question is required. This has been implemented for a twofold reason. First, to diminish the computational burden of the procedure. Second, to enable them to go streight to the point, and consequently facilitate the work of the organizing agent. 

\paragraph{Survey questions generation}
In this study, closed-ended questions were not designed primarily to achieve consensus, but rather to assess respondents' opinions and guide the discussion in insightful directions. 

\paragraph{Survey questions erogation}
Participants were asked to rate their responses on a scale of 1 to 5 for simplicity, as this range, compared to the traditional 1 to 7 scale, forces stronger decisions by reducing intermediate options. This approach was designed to capture clearer positions and facilitate a more structured analysis of expert perspectives.

\paragraph{Multiple rounds}
The Delphi process was iterated over five rounds, although no strict upper limit was inherently necessary. This constraint was imposed for experimental purposes, as literature on Delphi studies generally considers it a reasonable choice. Based on respondents' answers to closed-ended questions, new open-ended questions were generated. This iterative generation was crucial, as it allowed the discussion to move beyond the initial state and explore scenarios that would otherwise be difficult to conceive—ultimately fulfilling the core objective of a Delphi-based approach.

To address the issue of question proliferation, a workaround was implemented. Since both open-ended and closed-ended questions generated in each round tend to exceed the initial number, preventing an exponential increase was crucial to maintaining computational feasibility while preserving diversity. To achieve this, two filtering mechanisms were introduced. The first filter embeds all questions and removes those with a cosine similarity above a defined threshold. This step is essential because, by design, the organizing agent may generate closely related questions, especially when initial responses do not exhibit significant divergence. Removing these redundant questions reduces computational load without compromising the quality of the process. The second filter, also based on sentence embedding and cosine similarity, follows an iterative approach: it calculates the average similarity of each question with the others and sequentially eliminates the one with the highest mean similarity. This ensures that the final set of questions remains as diverse and dispersed as possible without leading to an explosion in their number. The number of retained open-ended and closed-ended questions is a tunable parameter of the model.

\paragraph{Summarize}
The validity of the results typically depends on three elelents. 
First, predictive validity is crucial, as the method's reliability is questioned if stable and verifiable predictions cannot be achieved \citep{helmer1983looking}. In our case, the Delphi study was repeated three times with the same configuration to assess if the results were analogous. 
Second, result validity is influenced by question design—while open-ended questions may lead to ambiguity and bias, closed-ended questions enhance clarity but may limit data richness \citep{keeney2011delphi}.
Finally, the validity of consensus \citep{keeney2011delphi}, which has not been considered in this case.

\subsection{Difference from classic Delphi methods}

The main differences between the classic Delphi method and the LLM-based proposed in this paper, beside the artificial nature of the experts involved, are the following.

First, the study does not stop when the consensus (or agreement \citep{meijering2013quantifying}) is achieved, but after a defined sets of rounds. This is the main difference between the traditional Delphi study and the LLM-based one proposed by this paper. According to its originators, the Delphi method was designed to achieve the highest possible reliability in expert consensus by engaging a group of specialists in a structured process involving multiple rounds of questionnaires with controlled feedback. This make sense when a human supervision is envolved \citep{dalkey1963experimental}. Our case, it was designed to not have the need to have an human supervision of the process, and because it was a purpose to get the most out of the different ideas without getting stucked. Of course, there could be always the chance pf putting an LLM-supervise that assess the consensus, but for the sake of experimentation and of proposing this also as a reasoning method for agents, we opted to have a defined limits in the number of rounds.

Second, given that there is not the problem of fatigue of the respondents, the open questions from the rounds after the first one are made by taking into consideration only the mean answers, and not selecting the most controversial answer with the standard deviation. 

Third, the fact of generating over-redundant questions, and removing them by means of sentence embedding and semantic similarity. Again, this could have been avoided, but we did it in order not to make this part susceptible to the recency bias and, in general, the poor ability of LLM to deal with very long texts. It would have been possible to give all the answers to the organizer and ask him to generate $n$ questions. This would not have been a good idea because it could have left some maybe elements. So we preferred to be redundant. 

Fourth, the absence of personal involvement helps mitigate certain issues. Specifically, the bandwagon effect can be considered negligible, as there are no differences in personality strength that could influence the results. Additionally, since LLMs respond unless explicitly instructed otherwise, there is no strong opposition between arguments that would naturally lead to a conciliatory position. Moreover, using LLMs eliminates the possibility of participant dropout during the process \citep{alarabiat2019delphi}.

\subsection{Experimental configuration}

Regarding the specific LLMs used as the agents' reasoning core, we conducted this experiment using a single LLM, specifically gpt-4o-mini-2024-07-18. This choice was made for two main reasons. First, we aimed to use an LLM accessible via OpenAI's API, as it is progressively becoming the industry standard. Second, this model offers an excellent balance between performance and cost, making it a suitable choice for the study.

The temperature of the LLMs was set to 0.7, which is the default value for OpenAI models. While this parameter alone would warrant a dedicated study to assess how choice dispersion affects the final outcome, it was kept at this level because this research represents a preliminary experiment. The goal of this article is not to explore all possible hyperparameter configurations of the LLMs or the LLM-based Delphi study and analyze their relationships. Instead, the focus is on presenting a new methodology and demonstrating its viability, aiming to encourage further experimentation within the research community.

In any case, a temperature greater than zero was deliberately chosen, ensuring that the system's agent components exhibit stochastic behavior. As a result, we expect to obtain different outcomes in each run, at least in the details. Additionally, there is an inherent stochasticity in the initial generation of experts. To account for this variability and assess the uncertainty of the results, the simulations were repeated three times.

\subsection{Implementation}

To implement the model, there were two strategies available. The first is to use LLMs based on a local machine, a server, or anyway any place where they could be uploaded and used by means of an API. The second instead was to access by an LLM by an external API available by another service. This second path was followed, for a twofold reason. First, it allows the access on the best model available, such as the one of OpenAI or Anthropic. Second, by means of parallelizing the access to the API, it allows to increase the speed of the entire process, which otherwise would be computationally expensive and time consuming. In any case, a budget is necessary for the computaitonally part, but not an high one.

The implementation was carried out using Python version 3.13.1, supplemented by several specialized libraries to enhance functionality. Since the main computational effort is handled by the server accessed via the API, the methodology does not require particularly powerful local computing resources. A commercially available personal computer is sufficient to run the simulation within a reasonable timeframe.

The whole system, together with the list of the libraries needed to make it run, is available e at the following repository: https://github.com/francescobertolotti/LLM\_Delphi.

\section{Results}

\begin{table}
\centering

\begin{tabular}{l|l|l|l|l|p{0.7cm}|p{0.6cm}|l|p{0.7cm}|l|l|l|l|l|l|l|l} 
ex & a & q & Geo. & Col. & Ec. Disp. & Ec. Eff. & Eth. & Tec. & Ed. & Reg. & Ind. & Cul. & Pri. & Psy. & Data & Sus. \\
\hline
1 & 5 & 5 & yes & yes & yes & no & yes & yes & yes & yes & no & no & no & no & no & no \\
2 & 5 & 5 & yes & yes & yes & no & yes & yes & yes & no & yes & no & no & no & no & no \\
3 & 5 & 5 & yes & yes & yes & no & yes & yes & yes & yes & yes & no & no & no & no & no \\
4 & 5 & 15 & yes & yes & yes & no & yes & yes & no & yes & no & yes & yes & no & no & no \\
5 & 5 & 15 & yes & yes & yes & no & yes & yes & yes & yes & yes & no & no & no & no & no \\
6 & 5 & 15 & yes & yes & yes & no & yes & yes & no & yes & no & no & no & no & yes & no \\
7 & 15 & 5 & yes & yes & yes & no & no & yes & yes & yes & yes & yes & no & no & yes & no \\
8 & 15 & 5 & yes & yes & yes & no & yes & no & yes & no & no & yes & no & yes & yes & yes \\
9 & 15 & 5 & yes & yes & yes & yes & yes & yes & no & yes & no & yes & no & no & yes & yes \\
10 & 15 & 15 & yes & yes & no & yes & no & yes & no & yes & no & no & yes & no & no & no \\
11 & 15 & 15 & yes & yes & yes & no & no & yes & no & yes & no & no & no & yes & no & no \\
12 & 15 & 15 & yes & yes & yes & no & yes & no & no & yes & no & no & no & no & no & yes \\
\hline
\end{tabular}
\caption{Summary of the conducted experiments, with the number of responding agents $a$ and the number of responses per agent $q$. Occurrences of each topic in each Delphi study.}
\end{table}

The analysis of results focuses on analyzing the final discussion and summary of results to obtain a broad understanding of the process. It is important to emphasize that the second type of analysis remains relevant, as when the process is used as an architecture for reasoning, the final user is primarily interested in the summary rather than the internal dynamics of the system. For the sake of clearness and readibility, each emerged topic is confined in a subsection. Finally, the novel points and the recurrent issues are highlighted. 

\subsection{Geopolitical issue}

Geopolitical tensions influence the strategic decisions of nations regarding artificial intelligence development and deployment. These tensions drive countries to prioritize self-sufficiency (experiment 1, experiment 4, experiment 6, experiment 8, experiment 9, experiment 11, experiment 12) over international cooperation, leading to potential fragmentation (experiment 1, experiment 3, experiment 4, experiment 7, experiment 9, experiment 10, experiment 12) and the emergence of rival technological ecosystems (experiment 9, experiment 10). This fragmentation can result in divergent technological standards (experiment 1, experiment 3, experiment 5, experiment 7) and pose a threat to cohesive innovation (experiment 3, experiment 6).

Such developments have been characterized through various terms, including “AI-Silos” (experiment 5, experiment 6), technological silos (experiment 10), “digital iron curtains” (experiment 8), and the onset of a “tech cold war” (experiment 10, experiment 12). This environment fosters the existence of distinct GenAI models and training datasets (experiment 6), further entrenching technological divides.

In parallel, national security concerns drive countries to address GenAI-related risks independently (experiment 2, experiment 11), even though such risks are inherently systemic and require coordinated mitigation efforts (experiment 2). Additionally, the localization of GenAI models to reflect specific cultural values may lead to a heterogeneous landscape of GenAI capabilities, creating a fragmented technological environment (experiment 8).

\subsection{Collaboration}

Collaboration is a fundamental driver of progress in GenAI development, as demonstrated by multiple experiments (experiment 1, experiment 2, experiment 3, experiment 4, experiment 11). However, geopolitical factors often lead nations to prioritize self-sufficiency over cooperation, increasing the risk of fragmentation (experiment 2, experiment 4, experiment 12).

Cooperation is essential as it enables several key advantages. It facilitates the establishment of a common innovation framework that fosters GenAI development (experiment 1) and supports responsible GenAI governance (experiment 5, experiment 10). Additionally, it promotes decentralized control, which helps address bias (experiment 1), integrate diverse cultural values (experiment 2), and ensure participatory design and oversight (experiment 11). Furthermore, cooperation supports the funding and development of open-source frameworks, enhancing accessibility and transparency in GenAI advancements (experiment 1, experiment 7).

Innovative partnerships should emerge as a response to these challenges (experiment 1, experiment 3), taking multiple forms, including regional collaboration (experiment 3), global AI consortia (experiment 1, experiment 3, experiment 9), and cross-border innovation hubs (experiment 1). These efforts contribute to the concept of a Global AI Commons (experiment 9), supporting equitable standards that ensure GenAI serves humanity rather than reinforcing the dominance of powerful nations (experiment 12).

A potential GenAI Governance Framework could integrate mechanisms such as immutable audit trails for transparency, decentralized identity certification for trust, smart contracts for automatic compliance, and tokenized accountability to incentivize ethical behavior among AI stakeholders (experiment 10).

To facilitate collaboration while navigating varying regulatory environments, AI Trade Zones could be established, allowing nations to operate under shared standards (experiment 6, experiment 7). Similarly, AI embassies could serve as neutral platforms for fostering international collaboration, setting ethical standards, and supporting joint research initiatives (experiment 10). These embassies could also mediate disputes, enhance capacity-building efforts, and promote cultural exchange (experiment 10).

A Decentralized AI Collaboration Network could further connect isolated innovation hubs, utilizing tokenized incentive systems, interoperable frameworks, decentralized governance models, and AI-driven collaboration tools to enable seamless knowledge exchange (experiment 12).

Despite these promising solutions, several obstacles hinder effective collaboration, beyond geopolitical concerns. These include digital divides, cultural resistance, interoperability issues, and algorithmic bias (experiment 7). Potential solutions to these challenges involve blockchain for secure data sharing, community-driven tech hubs, GenAI-powered communication tools, and open-source frameworks to promote inclusivity and cooperation (experiment 7).

\subsection{Economic disparities}

Economic disparities play a crucial role in shaping GenAI development, potentially exacerbating global inequalities if there is no common ground for investment (experiment 1, experiment 2). The pace of GenAI advancements is significantly influenced by economic conditions (experiment 6, experiment 12), which may widen the gap between wealthier and poorer nations.

To address this challenge, an AI Development Fund could be established, focusing on equity-based funding, local innovation hubs, and strategic partnerships to ensure that GenAI solutions generate economic benefits for local communities (experiment 9). Without such initiatives, the divide may deepen, particularly if open-source GenAI alternatives consistently underperform compared to proprietary solutions (experiment 6). Since poorer nations are more likely to adopt open-source technologies (experiment 6, experiment 9, experiment 12), disparities in quality and access could reinforce existing economic inequalities.

In this scenario, wealthier countries could continue to innovate at a rapid pace, while economically unstable regions may rely on grassroots GenAI movements to address local needs (experiment 3, experiment 4, experiment 7, experiment 9). A key factor contributing to this digital divide is the availability of pre-existing infrastructure, which can significantly influence a region’s ability to implement and scale GenAI solutions (experiment 4).

If economic disparities also translate into GenAI disparities, a potential consequence could be the emergence of "AI refugees"—talented individuals who migrate toward regions offering better technological support and funding opportunities (experiment 11). Addressing this issue requires leveraging local solutions that challenge dominant GenAI development paradigms (experiment 3) or that directly meet specific local needs (experiment 9).

Despite these challenges, there is also optimism that emerging economies can harness GenAI for specialized applications, helping create a more equitable technological landscape (experiment 4). This is especially feasible if GenAI governance remains localized and tailored to the needs of the communities it serves (experiment 9).

More broadly, economic conditions have a direct impact on the development of GenAI (experiment 5). Given that GenAI research and deployment require substantial resources, economic downturns may shift priorities away from ambitious GenAI projects toward addressing immediate social and economic challenges (experiment 5). To ensure a balanced approach, a GenAI impact evaluation framework should be developed, incorporating an assessment of its economic effects as a fundamental component (experiment 9).

\subsection{Economic effect}

Economic conditions play a crucial role in shaping the development and deployment of GenAI (experiment 10). During periods of economic recession, companies may prioritize cost-effective GenAI solutions that enhance efficiency and reduce expenses, favoring practical and immediately applicable technologies over long-term, high-risk innovations (experiment 10). Conversely, in times of economic prosperity, there is a greater willingness to invest in cutting-edge GenAI projects, fostering advancements in research and development and enabling the exploration of novel applications (experiment 10).

\subsection{Ethic}

Ethics plays a fundamental role in both the governance of GenAI and the way it is experimentally developed, particularly given the risks associated with GenAI deployment (experiment 5). To ensure ethical experimentation, regulatory ethics sandboxes (experiment 1, experiment 6) and real-time impact assessments (experiment 1, experiment 6) can be employed. These mechanisms provide controlled environments where GenAI technologies can be tested while continuously monitoring their societal and ethical implications.

Such real-time assessment processes could be supported by a broader ethical framework (experiment 2, experiment 3, experiment 6), which has been conceptualized as an “AI Treaty” (experiment 6). Unlike static regulatory models, this framework is designed to be dynamic, allowing it to adapt to rapid technological and market changes (experiment 2, experiment 3). The framework’s adaptability could be maintained through an evolutionary approach based on user feedback (experiment 4), ensuring that ethical considerations evolve alongside GenAI advancements.

To further structure GenAI governance, tools such as an Ethical Impact Scorecard could be implemented, along with additional metrics to address key concerns, including data sovereignty, regulatory differences, and intellectual property rights (experiment 12). However, for any ethical governance framework to be effective, it must achieve broad acceptance, ideally at a universal level (experiment 3) or at least through international consensus (experiment 4).

\subsection{Technological innovation}

GenAI development could be significantly advanced through AI-Co-Creation Platforms, which facilitate collaborative model building and governance (experiment 5). These platforms enable the incorporation of feedback loops during GenAI development, enhancing adaptability and refinement (experiment 5, experiment 10). Additionally, they promote decentralized governance, ensuring that GenAI models evolve in an open, transparent, and distributed manner (experiment 5).

A key enhancement for GenAI systems is contextual memory, which improves their ability to interact with dynamic environments (experiment 1), particularly when embedded into agents (experiment 3). The ability to adapt to different contexts is crucial for GenAI systems operating in diverse real-world settings (experiment 4). One proposed approach is the introduction of a dual-layer hierarchical memory system, allowing GenAI to distinguish between short-term situational context and long-term user preferences (experiment 9).

To ensure inclusivity and cultural relevance in GenAI systems, developments should focus on real-time bias detection in real-world environments, particularly in LLM agent-based applications (experiment 1). Furthermore, Cultural Contextual Assistants could be developed to enhance GenAI interactions by integrating culturally aware responses and behaviors (experiment 1).

LLM agents will be equipped with self-improvement mechanisms (experiment 3) and multi-modal sensory feedback systems (experiment 3), enabling them to perceive and respond to the world in a more dynamic manner. This evolution will position GenAI-based agents as GenAI companions with a pervasive role in daily life (experiment 3).

Real-time learning is essential not only for GenAI agents but also for GenAI systems in general (experiment 6). This is particularly relevant for applications that require user consent protocols, especially in cases involving emotional data processing (experiment 11). Beyond real-time adaptation, GenAI learning could also be federated, enhancing data privacy management (experiment 10) while ensuring emotional authenticity and ethical transparency (experiment 11).

The integration of Neuro-Symbolic AI is considered a promising pathway for improving reasoning capabilities (experiment 2). Its relevance could increase when implemented within modular architectures (experiment 3, experiment 6), particularly transformers (experiment 11). In this context, the development of a Symbolic-Transformer Interface has been proposed to reconcile symbolic reasoning with statistical learning, addressing inefficiencies in processing and understanding complex concepts (experiment 5).

More broadly, modular architectures represent an important future direction for GenAI (experiment 3, experiment 4, experiment 6), as they enable scalability and interoperability (experiment 4). Interoperability could be further enhanced through the establishment of a marketplace for AI modules (experiment 4) or even for entire models (experiment 10).

Participants also proposed adaptive attention mechanisms and dynamic architectures that allow models to adjust their focus in response to specific tasks and contexts (experiment 2). Beyond transformers, alternative architectures like Graph-Guided Generative Adversarial Networks could play a pivotal role in GenAI development (experiment 5). Graph-Guided Generative Adversarial Networks are particularly relevant for fields such as drug discovery, urban planning, social network analysis, and personalized education (experiment 5), as they offer superior capabilities in processing complex data structures (experiment 5). The integration of Graph-Guided Generative Adversarial Networks could mark a paradigm shift in how GenAI models address and solve highly intricate problems across diverse domains.

Multi-modal fusion is expected to play a significant role in the evolution of GenAI systems (experiment 2). This advancement could enhance the ability of GenAI companions to interact with and influence their environments (experiment 3) while also expanding the storytelling potential of GenAI (experiment 7).

To integrate ethics into GenAI systems, a decentralized platform utilizing blockchain technology could be developed to certify and trade ethical GenAI frameworks (experiment 2). Such a system would ensure transparency and trust in GenAI governance, enabling accountability while fostering collaboration among diverse stakeholders.

A related element is the necessity for greater explainability in GenAI (experiment 4, experiment 5, experiment 6). This could be achieved through further customization of GenAI models (experiment 4), the integration of narrative approaches into explanations (experiment 4, experiment 10), or the use of dynamic knowledge graphs to make GenAI reasoning more transparent. A fundamental challenge is balancing innovation with explainability (experiment 6), ensuring that GenAI remains both powerful and interpretable.

In a related effort, the integration of emotional intelligence into GenAI is increasingly recognized as essential (experiment 6). This is particularly important to ensure that GenAI-generated outputs align with societal values (experiment 6). Emotional intelligence could be enhanced through reasoning processes that explicitly incorporate ethical considerations (experiment 6). Furthermore, GenAI systems should anticipate user needs and emotions, leading to more empathetic and engaging interactions (experiment 9).

A proposed "Story Weave" platform would enable users to co-create narratives in real-time, balancing personalization with communal input to maintain diverse perspectives while still catering to individual preferences (experiment 7). This platform should include real-time collaborative storytelling tools, allowing users to seamlessly contribute to narrative threads (experiment 7). It highlights the potential for dynamic character evolution, multimodal inputs, real-time collaboration, and ethical storytelling frameworks (experiment 7).

The scalability and interoperability of GenAI systems are also crucial factors. Increasing scalability and interoperability could lead to more effective integration of diverse data sources and improve GenAI adaptability across different domains (experiment 4).

These developments require significant computational resources, and one way to address these demands is through edge computing, which could help manage data processing more efficiently by distributing computational loads closer to the data source (experiment 10).

Finally, fostering community engagement in GenAI design is essential. Actively involving local residents in the GenAI development process ensures that their insights and cultural perspectives are directly incorporated, leading to more inclusive and socially aware GenAI systems (experiment 8).

\subsection{Education}

AI literacy is a fundamental aspect of technological development and social integration (experiment 1, experiment 2), particularly in underserved communities where access to education and digital resources is often limited (experiment 1). Addressing this issue requires targeted strategies that consider different demographic groups (experiment 2) to ensure inclusivity and effectiveness. The primary objectives of AI literacy initiatives should be to enhance critical thinking (experiment 2) and foster ethical reasoning, enabling individuals to engage thoughtfully with GenAI systems and understand their broader implications (experiment 2).

To measure the effectiveness of such initiatives and track public engagement with GenAI technologies, the implementation of an AI Literacy Index could provide valuable insights, serving e as a tool for assessing awareness, comprehension, and responsible use of GenAI across diverse populations, helping policymakers and educators design more impactful GenAI education programs (experiment 8).

\subsection{Regulation}

Any regulatory framework for GenAI should be either decentralized and context-specific (experiment 4, experiment 8) or globally comprehensive to ensure consistency and fairness across different jurisdictions (experiment 8). Given the rapid pace of technological advancements, a dynamic regulatory framework that evolves in real time alongside GenAI technologies is necessary (experiment 1, experiment 8). 

A key emerging concept is AI diplomacy, which could play a pivotal role in fostering international agreements on GenAI regulations and ethical standards (experiment 1, experiment 12). Additionally, integrating Real-Time Sentiment Analysis into GenAI governance structures could enable a more democratic and responsive approach, allowing regulatory bodies to adapt policies based on public perception and concerns regarding GenAI (experiment 2).

Regulatory frameworks must strike a balance between respecting local values and promoting international collaboration (experiment 4). If regulations remain fragmented at the national level rather than being globally harmonized, disparities in GenAI development could widen, as different policies may either hinder or accelerate GenAI progress depending on the region (experiment 5). To bridge this gap, regulatory initiatives should be developed by interdisciplinary teams that incorporate expertise from diverse backgrounds, ensuring that policies align with both technological advancements and cultural sensitivities (experiment 8, experiment 9).

An effective regulatory strategy must also balance strict oversight with innovation. While stricter regulations can enhance responsible GenAI development, overly permissive frameworks risk enabling uncontrolled technological advancements that may pose ethical and societal risks (experiment 10).

A Universal AI Protocol could provide a structured regulatory foundation, incorporating an Ethical Impact Scorecard and additional metrics to address data sovereignty, regulatory differences, and intellectual property concerns (experiment 12). 

\subsection{Industries affected}

\begin{table}
\centering

\begin{tabular}{p{0.8cm}|p{1.4cm}|p{1.4cm}|p{1.4cm}|p{1.4cm}|p{1.3cm}|p{1.5cm}|p{1.5cm}|p{1.2cm}} 
\hline
exp & Healthcare & Creativity & Education & Urban planning & Disaster & Agriculture & Translation & Social network \\
\hline
1 & no & no & no & no & no & no & no & no \\
2 & yes & yes & yes & yes & no & no & no & no \\
3 & yes & no & yes & no & yes & no & no & no \\
4 & no & no & no & no & no & no & no & no \\
5 & yes & yes & yes & yes & yes & yes & yes & yes \\
6 & no & no & no & no & no & no & no & no \\
7 & no & yes & no & no & no & no & no & no \\
8 & no & no & no & no & no & no & no & no \\
9 & no & no & no & no & no & no & no & no \\
10 & no & no & no & no & no & no & no & no \\
11 & no & no & no & no & no & no & no & no \\
12 & no & no & no & no & no & no & no & no \\
\hline
\end{tabular}

\end{table}

GenAI has the potential to transform a wide range of industries, driving innovation and enhancing efficiency across multiple domains. In medicine, GenAI can enable more personalized treatments by analyzing patient data to tailor medical interventions to individual needs (experiment 2). Additionally, it can play a crucial role in drug discovery, accelerating the identification of new compounds and optimizing pharmaceutical research (experiment 5).

In creative industries, GenAI systems can function as co-creators, augmenting human creativity by generating novel ideas and artistic expressions (experiment 2, experiment 5). The integration of GenAI for narrative feedback and gamification can further enhance user engagement and creativity in storytelling, allowing for more immersive and interactive experiences (experiment 7).

The education sector can benefit significantly from GenAI, particularly in adaptive learning systems that adjust to individual student needs, improving knowledge retention and engagement (experiment 2). The introduction of personalized GenAI companions could further support students by providing customized tutoring and guidance, making education more accessible and tailored to different learning styles (experiment 3). More broadly, personalized education frameworks supported by GenAI can lead to a more efficient and inclusive learning environment (experiment 5).

Urban planning is another field where GenAI can have a substantial impact. GenAI models can optimize urban design by analyzing spatial patterns, infrastructure efficiency, and sustainability factors to create more livable cities (experiment 2). Advanced simulation capabilities allow for the visualization of entire urban landscapes before construction, helping to optimize infrastructure, public spaces, and environmental impact (experiment 5).

In disaster response, GenAI can improve situational awareness and decision-making. Personalized GenAI companions can assist emergency responders by analyzing real-time data to provide situational guidance (experiment 3). Additionally, intelligent robots powered by GenAI can assess environments and autonomously make decisions in hazardous conditions where human intervention is too risky (experiment 5, experiment 9).

GenAI also has significant implications for social network analysis. The integration of GenAI-driven systems can synthesize realistic social interactions and user behavior patterns, enhancing recommendation algorithms and enabling businesses to predict trends with greater accuracy. This allows companies to dynamically tailor their strategies based on evolving social behaviors (experiment 5).

Beyond these specific industries, the overall advancement of the GenAI field could be driven by localized innovation, where GenAI solutions are developed to address specific regional and community needs (experiment 5). 

\subsection{Culture}

Emerging economies have the potential to leverage GenAI for culturally specific applications, promoting a more equitable technological landscape (experiment 4). To foster collaboration across cultures and nations, initiatives such as the Cultural Ethics Charter could play a significant role, establishing shared ethical guidelines for GenAI development that respect diverse cultural norms and traditions while facilitating international cooperation in GenAI governance (experiment 4).

From a cultural perspective, progressive customization of GenAI systems presents a promising approach to increasing trust and acceptance (experiment 4). By allowing users to adapt GenAI tools to their specific cultural contexts, these systems can become more relatable and widely embraced. Similarly, the development of culturally adapted GenAI tools designed to resonate with local communities can enhance both the effectiveness and societal acceptance of GenAI technologies (experiment 7).

The creation of community-driven innovation hubs could further ensure that GenAI development reflects local cultural values (experiment 8, experiment 9). These hubs would serve as collaborative spaces where GenAI models are designed with input from local stakeholders, fostering inclusive technological progress that aligns with community needs (experiment 9).

To assess the broader impact of GenAI on cultural identity and production, metrics should be developed to evaluate how GenAI influences cultural representation and preservation (experiment 8). Such evaluations should not only measure the ongoing effects of GenAI but also be conducted ex-ante, before deploying GenAI systems, to ensure that they align with community values and ethical standards (experiment 8). Additionally, GenAI developers could be trained in local customs and cultural values to enhance the design of systems that are both respectful and adaptive to diverse cultural landscapes (experiment 8).

\subsection{Privacy}

The implementation of contextual adaptability in GenAI systems has raised significant concerns regarding privacy and user autonomy, since as GenAI models become more adaptive to individual and regional contexts, the risk of excessive data collection and potential misuse increases, making it crucial to establish robust privacy safeguards to protect users' rights and ensure transparency in data usage (experiment 4).

Data privacy is also a critical issue in the context of global collaboration, where cross-border GenAI initiatives require mechanisms to ensure that sensitive information remains secure while enabling international cooperation (experiment 7). A possible solution to mitigate these risks is the implementation of federated learning, which allows GenAI models to be trained across multiple decentralized data sources without the need to transfer raw data (experiment 10). 

\subsection{Psychology and society}

The psychological effects of GenAI interactions should be systematically monitored to promote overall well-being (experiment 8, experiment 11), particularly considering their potential long-term impacts (experiment 11). 

To address the broader societal implications of GenAI, a Societal Impact Assessment framework should be developed to evaluate GenAI technologies at various stages of their development (experiment 8). Alternatively, societal impact assessments should be integrated into a more comprehensive governance framework that considers ethical, economic, and cultural dimensions (experiment 9). These assessments should be conducted in real-time, leveraging social media and community platforms to continuously gather public feedback and assess the evolving societal effects of GenAI technologies (experiment 8).

To ensure that GenAI aligns with societal values, several metrics and evaluation tools have been proposed, including an Ethical Impact Score, a Transparency Ledger, Inclusivity Metrics, and an Adaptability Quotient: these instruments could provide structured assessments of GenAI fairness, accessibility, and adaptability across different demographic and cultural groups (experiment 11). 

Additionally, securing GenAI-generated content is a pressing concern. Possible solutions include blockchain technology to verify content origin, innovative watermarking techniques to track authenticity, and anomaly detection systems to monitor and regulate GenAI-generated content across digital platforms (experiment 10). Finally, to enhance emotional well-being and inclusivity in GenAI-generated environments, novel concepts such as customizable safe spaces, dynamic emotional landscapes, and community engagement mechanisms could be implemented (experiment 11). 

\subsection{Data}

Data privacy presents a significant challenge for global GenAI collaboration (experiment 7). While nations have attempted to mitigate these concerns by developing AI-silos, this approach comes at the cost of restricting access to diverse datasets, which are crucial for improving GenAI performance and reducing biases (experiment 8). 

One potential solution to balance data privacy and accessibility is the development of dedicated blockchains (experiment 7). Additionally, policy frameworks should be established to empower communities in governing the data that represents their culture, ensuring that GenAI systems are developed and deployed in an ethically responsible manner (experiment 8).

\subsection{Sustainability}

Recognizing the ecological impact of GenAI development is essential, and advocating for sustainable practices in its design and deployment is a key priority (experiment 8). The increasing computational demands of GenAI models contribute to energy consumption and environmental costs, necessitating strategies to minimize their ecological footprint. To address this challenge, it is crucial to implement assessment mechanisms that evaluate the environmental impact of GenAI systems throughout their lifecycle (experiment 12). A GenAI impact evaluation framework should be established, incorporating sustainability assessments as a fundamental component (experiment 9). 

\section{Discussion}

The discussion of the results has been structured into two main sections. 
The first section provides a detailed analysis of the novel LLM-based Delphi method, clarifying its strengths and limitations. Given that this study represents a preliminary effort, the primary objective is to lay the groundwork for future improvements in this domain. The discussion emphasizes how this method refines traditional Delphi approaches through large language models, while also identifying potential challenges and limitations.
In the second section, the results generated by the LLM-based Delphi study are examined and summarized. This part focuses on interpreting the findings, contextualizing them within the broader scope of the research, and evaluating their implications. 

\subsection{LLM-based Delphi}

One of the well-documented challenges in traditional Delphi studies is the potential for participant bias, which can influence the consensus-building process. In this study, an attempt was made to mitigate bias by generating a diverse set of GenAI agents with varied backgrounds, ensuring a broader range of perspectives. While this approach presents a promising direction, it also raises important questions regarding the extent to which LLM behavior can be modified and customized solely through System Prompt adaptation. The underlying concern is that biases inherent in the pre-training phase of the model may still manifest, despite efforts to diversify perspectives at the prompt level. This issue extends to a broader philosophical and technical debate: whether any cognitive entity can function without bias and whether any trained system is inevitably shaped by the data it was exposed to. From this perspective, bias might be an intrinsic property of both human and artificial decision-making processes, rather than a defect that can be entirely eliminated. An alternative approach could involve fine-tuning LLMs specifically for Delphi-based applications, explicitly introducing controlled biases that align with the study’s goals. While this strategy might allow for greater transparency and intentionality in bias management, it would also require significant computational resources and financial investment, making it a potentially cost-prohibitive solution.

A fundamental challenge in utilizing LLMs for Delphi-based studies is the lack of precise knowledge about their pre-existing information, which, while inferable through certain techniques, remains largely opaque. Given that this study aims to test a generalizable methodology, it was considered preferable to experiment directly rather than preemptively explore the model’s knowledge on specific topics. This decision aligns with the broader objective of evaluating the adaptability of LLMs in structured deliberation processes without being constrained by predefined assumptions about their informational scope. Since LLM knowledge cannot be fully known but only inferred, one possible approach is to start with a broad set of instructions and later refine the focus in subsequent phases of experimentation. This iterative method allows for adaptive control over the study while acknowledging the inherent uncertainty in the LLM’s informational baseline. Fine-tuning the model to embed specific knowledge could improve its reliability and contextual understanding, but this remains a resource-intensive and costly process that may not always be feasible. An alternative approach involves the implementation of Retrieval-Augmented Generation systems, where the model retrieves relevant external information before generating responses. This strategy allows the information sources to be explicitly defined and assessed, mitigating the risks of relying on unknown or potentially outdated internal knowledge. To evaluate the impact of such an approach, experiments should be conducted to analyze how responses differ when agents have access to different sets of information. This would introduce an additional layer of heterogeneity, not only in the roles assigned to different agents but also in the knowledge available to them, potentially enhancing the depth and diversity of deliberation in the LLM-based Delphi process.

Also, LLMs do not necessarily possess expert-level knowledge across all domains, and their performance can vary significantly depending on the topic. This limitation became particularly evident when the model addressed Graph-Guided Generative Adversarial Networks, where some degree of conceptual confusion was observed. While the system was capable of generating responses and engaging with the topic, the results indicated inconsistencies and gaps in its understanding, suggesting that its knowledge in this area was either incomplete or misaligned with current research.

Providing external information that extends beyond the training set of the LLM agents could help mitigate the impact of knowledge cutoffs and their effect on prediction accuracy. In the experiments conducted for this study, data was generated in February 2025, while the model used, GPT-4o-mini, had a knowledge cutoff in October 2023. This implies that the model was making predictions based on information that was already 15 months outdated at the time of the study. While in some domains this temporal gap may not significantly impact the quality of responses, in fields that evolve rapidly—such as GenAI—this lack of up-to-date knowledge poses a substantial limitation. In these fast-moving areas, the model’s starting assumptions may already be outdated, leading to conclusions that fail to reflect recent advancements. This highlights the necessity of providing external, real-time information to ensure that GenAI-assisted methodologies remain relevant and accurate. Without mechanisms to supplement or update knowledge, predictions and insights generated by the model risk being constrained by historical limitations, reducing their applicability in dynamic research areas.

Unlike traditional human-based Delphi studies, the use of LLM agents eliminates the issue of respondent fatigue. In human-driven processes, it happens that participants shorten their responses or even withdraw from the study to conclude it more quickly. This limitation can reduce the depth of deliberation and the overall quality of insights generated. In contrast, LLM agents do not experience such constraints, allowing for the inclusion of a greater number of open-ended questions without the risk of diminishing engagement. Additionally, the methodology can support more iterative rounds, refining the responses through successive interactions in a way that would be difficult to achieve with human participants. This advantage enables a more comprehensive exploration of the subject matter, as the study is not constrained by the cognitive and time-related limitations of human respondents. 

Taking a broader perspective and considering this approach as a multi-agent reasoning architecture, encapsulating the Delphi study process within a stochastic function allows for an analysis of the uncertainty in the results. This aspect is inherently impossible in traditional human-based Delphi studies, as the act of participating and reaching a consensus alters the participants themselves, making it impossible to replicate the same study twice identically. In contrast, in an LLM-based Delphi framework, the process can be repeated under controlled conditions, provided that all hyperparameters, particularly the temperature of the models, remain greater than zero to introduce an element of stochasticity that can be analyzed or at least considered in the interpretation of results.
To assess this variability, each combination of questions and the number of agents was repeated three times during the experiments. This approach allowed us to observe how frequently the same responses were generated and how much variability emerged even under identical configurations. The observed variability, as presented in the results tables, can be attributed to two factors: the intrinsic stochastic nature of individual LLM responses and the propagation of these variations throughout the system. Given the star-like structure of this architecture, in which responding agents, even if indirectly, interact with each other, their repeated interactions lead to emergent unpredictability beyond the initial conditions.
An interesting avenue for future analysis involves investigating the sensitivity of the study to initial conditions. Even when setting temperature to zero, certain variables could significantly influence the results, such as the specific system prompt, the formulation and order of the initial questions, and the hyperparameters of the model or even the choice of models themselves. This sensitivity assessment could provide deeper insights into how stable and reproducible Delphi-based methodologies are, extending beyond LLM-based approaches to evaluate the robustness of traditional Delphi studies. By considering participants and organizers as a single system, the results obtained here offer a new lens through which to examine the general reliability and consistency of Delphi studies in the existing literature.

One notable observation regarding the sensitivity to initial conditions concerns, which emerges from the results, is the order and recurrence of certain elements in the responses. Specifically, the first elements that emerged were consistently related to economic disparities between nations or geopolitical tensions. While other recurring themes were present in the responses, as shown in the results tables, these particular topics always appeared at the beginning of the generated discussions. This raises an important question regarding whether this pattern is a consequence of the specific formulation of the initial questions or if it reflects intrinsic preferences within the LLMs themselves.
The chaotic nature of this system is not necessarily a limitation when the goal is to generate new predictions and explore potential future scenarios, as it allows for a diverse and dynamic range of responses. However, this unpredictability could pose challenges if the system is intended to be used as a stable reasoning tool. In such cases, ensuring consistency and reducing sensitivity to initial conditions would be crucial. Understanding whether these emergent patterns are driven by prompt design or model tendencies would provide valuable insights into the behavior and reliability of LLM-based Delphi methodologies and their applicability in different contexts.

Some aspects that emerged in the results appear to be direct consequences of the questions posed, while others arise in a less predictable manner. Certain themes, such as the focus on education or references to privacy, do not seem to have a strong or immediate relationship with the initial prompts, making it unclear whether they stem from latent patterns in the model's training data or from the structured interactions between agents. In some cases, entirely new concepts surfaced, suggesting that the system is not merely reflecting input biases but is also generating novel associations that extend beyond the explicit scope of the questions.
Further analysis is necessary to determine the extent to which these unexpected outcomes are systematic artifacts of LLM behavior or whether they indicate a deeper emergent property of structured interactions among agents capable of cognitively complex tasks. If certain ideas consistently emerge despite not being explicitly prompted, it may suggest that multi-agent reasoning frameworks, at a certain threshold of innovation and conceptual expansion, could be understood as an instance of emergence, where structured interactions among GenAI agents lead to unexpected yet coherent knowledge production that is not easily reducible to the individual components of the system.

The presence of multiple agents, even when they share the same underlying model, combined with the inherent stochasticity in their responses, allows for the generation of ideas over time, provided that interactions occur over a sufficiently long trajectory. While this process inevitably leads to a significant amount of repetition, it also creates opportunities for unexpected insights to surface. The system can be considered to be suited to address "black swan" problems, where the objective is not for each agent to generate new and valuable ideas at every iteration, but rather for a single instance of innovation to emerge at least once.
In this sense, the process does not require consistent creativity from every agent at every step, but rather the generation of at least one significant breakthrough, which, once surfaced, can be analyzed and integrated into broader discussions, and released in the output.

The computational cost of the model remains a significant constraint, even when API calls are parallelized, as the total execution time remains substantial. This limitation suggests that the approach is likely to be applicable only to highly relevant problems and specific well-suited issues where the benefits outweigh the computational expense. However, this challenge is not fundamentally different from what occurs in human cognitive processes. Humans do not generate groundbreaking ideas on a continuous basis; instead, innovative insights emerge sporadically, while much of the remaining cognitive effort is dedicated to implementing, refining, and evaluating these ideas.
In this sense, the computational burden of the model does not necessarily diminish its utility but rather mirrors the natural constraints of creative and analytical thought. Just as human problem-solving involves alternating between idea generation and execution, an GenAI-driven system may function most effectively when deployed in a similar manner—periodically producing novel insights while allocating significant resources to assessment and validation. 
 
Finally, it is useful for the reader to summarize the main limitations that have emerged during the discussion. They are the difficulties in adapting the behavior of LLMs to guarantee heterogeneity in knowledge and behavior, the presence of LLM bias, the presence of knowledge cutoffs, the coherence in terms of the sensitivity between the initial conditions and the resulting output, and the computational cost.

\subsection{Generated results}

Several key points emerged from the discussion. 
First, regulatory decisions must account for the rapidly evolving nature of the field and should be as widely shared as possible. A formal organization dedicated to overseeing these regulatory changes could help ensure consistency and adaptability in governance. 
Second, the impact of GenAI on various industries is still unfolding, even if the core technology remains unchanged, suggesting that the most significant transformations are yet to come.
Third, GenAI has the potential to create winners and losers across different levels of society, making it crucial to anticipate and address disparities in advance. 
Fourth, the effects of GenAI are interdependent, meaning that certain innovations, such as an GenAI module marketplace, would not be viable without common regulatory frameworks and shared infrastructure. 
Fifth, the need for exchange platforms is another recurring theme, whether they involve frameworks, GenAI modules, or broader collaboration mechanisms. Although such platforms are sometimes referred to as marketplaces, the discussion did not focus on their economic or competitive dimensions, instead emphasizing the importance of fostering cooperation.
Sixth, the study placed greater emphasis on the sociological aspects of GenAI rather than the technical ones, partly because this focus aligns with the strengths of large language models. 
Finally, a recurring idea is that to manage and control the societal and technological impacts of GenAI, robust measurement mechanisms are necessary. The ability to track and assess these changes systematically would be essential for informed decision-making and responsible GenAI governance.

A noticeable bias in the responses appears to be the predominance of positive perspectives on GenAI, with challenges and risks being mentioned only tangentially. While a range of potential issues, including social and sustainability concerns, did emerge, they were not central to the discussion. This outcome is partially influenced by the framing of the questions, which were designed with a constructive and forward-looking approach rather than a critical one. However, it is also evident that many of the identified problems were framed in terms of GenAI’s lack of diffusion, implicitly assuming that broader adoption is inherently beneficial—a perspective that is not necessarily justified.
Certain controversial or minority perspectives were almost entirely absent from the responses. Topics such as existential risks, the possibility of GenAI systems achieving consciousness, or the ethical implications of GenAI being perceived as sentient were not addressed at all. This absence raises questions about whether these omissions are due to intrinsic biases in the models, the nature of the dataset on which they were trained, or simply the way the study was structured. A more deliberate inclusion of critical perspectives in future iterations could help provide a more balanced and comprehensive exploration of the topic, ensuring that both the benefits and risks of GenAI development are adequately considered.

\section{Conclusions}

This study introduces a novel LLM-based Delphi methodology, designed both as a framework for conducting Delphi studies where LLMs replace human experts and as a potential reasoning architecture for GenAI systems. By leveraging LLMs to elicit domain-specific knowledge, this approach effectively mitigates traditional limitations of human-driven Delphi studies, such as bandwagon effects and respondent fatigue. Unlike the classic Delphi approach, which aims at achieving consensus, this method is specifically designed to generate ideas, emphasizing divergent thinking and exploration rather than agreement among participants.
The methodology was applied to the specific case of the future of GenAI, producing valuable insights both on the subject matter itself and on the LLM-based Delphi process as a structured reasoning tool. The study not only demonstrated the viability of this approach in generating structured discussions but also provided a deeper understanding of its mechanisms, limitations, and potential refinements. By testing this method in a real-world scenario, it was possible to assess its strengths and challenges, paving the way for future improvements and broader applications in GenAI-assisted decision-making and structured deliberation frameworks.

The objective of this work was not to exhaustively explore all possible variations of the LLM-based Delphi methodology, but rather to introduce it, conduct an initial experimental validation, and outline its main strengths and limitations. This study serves as a foundation for further research, providing a structured approach to assessing the feasibility and applicability of using LLMs as substitutes for human experts in Delphi-style studies.
Given the preliminary nature of this investigation, there remains significant research to be conducted to fully explore the potential of this methodology. Further studies are needed to refine the approach, optimize its effectiveness, and develop a deeper understanding of both its capabilities and constraints. The findings presented here offer an initial perspective on the strengths and weaknesses of this framework, highlighting areas where improvements can be made.

Future developments of this methodology could involve modifications to the current approach, such as removing closed-answer formats and analyzing the impact on idea generation. This adjustment would allow for a deeper exploration of how LLM agents generate novel insights when not constrained by predefined response structures. Another potential avenue of research involves introducing heterogeneity in the respondent agents by mixing different LLMs, examining whether variations in training data and model architectures increase the diversity of individual responses and lead to more creative final outputs at the system level.
A promising direction could be the development of a hybrid human-LLM Delphi technique, leveraging the strengths of both GenAI and human expertise to improve the quality and depth of responses. This integration could enable more nuanced deliberation, balancing the computational efficiency of LLMs with the critical thinking and contextual awareness of human participants. Additionally, one of the key challenges discussed in this study is the knowledge limitations of LLMs, particularly due to training cutoffs and lack of real-time updates. A potential solution is to integrate the system with a Retrieval-Augmented Generation framework, allowing access to external sources of information that could enrich responses. Providing different LLM agents with varied document sets could further increase the heterogeneity of perspectives, leading to a more diverse and comprehensive decision-making process.
Another essential aspect for future exploration is an analysis of the system’s sensitivity to various factors, including hyperparameters, prompt design, and the underlying knowledge base. Understanding how different configurations affect the stability, reliability, and variability of responses would be crucial for refining this methodology and ensuring its adaptability across different domains and applications.

\section{Aknowledgement}
The author F.B. would like to thank V. for the invaluable stimulus in assessing the possible future of GenAI and now experimenting with ways to use LLMs. It is unlikely that such an idea would have arisen without their valuable contribution. 

\bibliographystyle{plainnat}
\bibliography{references}  


\end{document}